\documentclass[11pt]{article}

\usepackage[letterpaper,margin=1in]{geometry}
\usepackage{booktabs}
\usepackage{array}
\usepackage{enumitem}
\usepackage{graphicx}
\usepackage{hyperref}
\usepackage{xcolor}
\usepackage{tikz}
\usetikzlibrary{arrows.meta,calc,fit,positioning}

\hypersetup{
  colorlinks=true,
  linkcolor=blue,
  citecolor=blue,
  urlcolor=blue
}

\title{ChromaFlow: A Negative Ablation Study of Orchestration Overhead in Tool-Augmented Agent Evaluation}
\author{Tarun Mittal\\Octave-X}
\date{May 18, 2026}

\begin{document}
\maketitle

\begin{abstract}
Autonomous language-model agents increasingly combine planning, tool use,
document processing, browsing, code execution, and verification loops. These
capabilities make agent systems more useful, but they also introduce operational
failure modes that are not visible from final accuracy alone. This report
presents ChromaFlow, a tool-augmented autonomous reasoning framework built
around planner-directed execution, specialized tool use, and telemetry-driven
evaluation. We analyze ChromaFlow on GAIA 2023 Level-1 validation tasks under
clean evaluation constraints. A frozen full Level-1 baseline achieved 29/53
correct answers, or 54.72\%. A later recovery configuration with expanded
orchestration achieved 27/53 correct answers, or 50.94\%, while increasing
tracebacks, timeout events, tool-failure mentions, token-line calls, and
campaign-log cost estimates. Two randomized 20-task smoke evaluations produced
12/20 and 11/20 correct answers, showing that small diagnostic gains can be
unstable across samples. The central result is therefore a negative ablation:
more aggressive orchestration did not improve full-set performance and increased
operational noise. A later strict-provider full-Level-1 diagnostic reached
30/53, or 56.60\%, under explicit integrity controls, but at substantially
higher token-log cost. The report argues that bounded planner escalation,
deterministic extraction, evidence reconciliation, provider-health gates, and
explicit run gates should be treated as first-order requirements for reliable
autonomous agent evaluation.
\end{abstract}

\noindent\textbf{Keywords:} autonomous agents, tool-augmented reasoning,
multi-agent systems, GAIA, benchmark evaluation, reliability, negative
ablation, provider controls

\section{Introduction}

Large language model systems have moved from single-turn text generation toward
autonomous execution loops that plan, call tools, inspect intermediate evidence,
and revise answers. Systems such as ReAct showed the value of interleaving
reasoning and actions, while tool-use work such as Toolformer demonstrated that
language models can benefit from external APIs and structured tool calls
\cite{yao2023react,schick2023toolformer}. Benchmark suites such as GAIA measure
longer-horizon assistant behavior that often requires search, file handling,
multimodal reasoning, and careful answer synthesis \cite{mialon2023gaia}.
Since GAIA, agent evaluation has broadened toward multi-environment agent
benchmarks, realistic web and software-engineering tasks, operating-system
interaction, enterprise workflows, tool-agent-user interaction, and explicit
orchestration benchmarks
\cite{liu2023agentbench,zhou2023webarena,jimenez2023swebench,xie2024osworld,drouin2024workarena,yao2024taubench,ahn2026orchestrationbench}.
Recent agent-evaluation and agent-skills work also emphasizes reliability,
contamination, evolution, governance, and failure modes beyond raw tool use
\cite{dong2026agency,jiang2026sok}. This paper is positioned within that
evaluation-methodology line: it studies when additional orchestration fails to
produce a reliable improvement.

These systems are frequently evaluated by final task accuracy. Accuracy is
necessary, but it is incomplete. A configuration can preserve or even improve
accuracy while becoming more expensive, slower, harder to reproduce, or more
dependent on brittle retry loops. Conversely, a negative result can still be
useful if it exposes which architectural choices increase execution instability.
This paper treats operational telemetry as part of the evaluation object rather
than as incidental logging.

ChromaFlow is an autonomous reasoning framework designed for planner-directed
tool use and benchmark execution. It supports document analysis, web search,
code execution, structured extraction, browser interaction, and final-answer
verification. The system was developed to study whether adaptive orchestration
and reliability hardening can improve benchmark performance under clean
evaluation constraints.

The main finding is deliberately conservative. A smaller 20-task diagnostic
signal suggested that targeted reliability fixes could help. However, a later
full Level-1 recovery run did not preserve the improvement. The recovery
configuration scored lower than the frozen baseline and generated more
operational noise. A subsequent randomized smoke run also failed the gate for a
new full run. We therefore present the initial result as a negative ablation
and reliability analysis, not as a leaderboard claim. We then report a later
strict-provider full-Level-1 diagnostic that recovered a small observed gain,
30/53 instead of 29/53, while increasing token-log cost substantially.

\section{Contributions}

This report makes four focused contributions:
\begin{itemize}[leftmargin=*]
  \item It documents a negative ablation of autonomous-agent orchestration in
        which a more aggressive recovery configuration reduced full GAIA
        Level-1 accuracy while increasing operational noise and cost estimates.
  \item It provides a clean-evaluation reporting protocol that preserves
        benchmark integrity by publishing aggregate metrics while withholding
        task text, task identifiers, gold answers, predictions, attachments,
        raw URLs, and raw traces.
  \item It proposes reliability gates for agent benchmarks: smoke-run gains
        should justify full runs only when they clear both accuracy and
        operational-noise thresholds under randomized sampling.
  \item It adds a follow-up strict-provider diagnostic showing an observed
        one-task full-Level-1 improvement under recorded controls, while
        emphasizing that the magnitude is too small to support a broad
        capability claim and that cost efficiency remains unresolved.
\end{itemize}

This paper does not claim state-of-the-art performance on GAIA. Instead, it
contributes a negative systems ablation and a later controlled recovery
diagnostic. The negative ablation shows that a more aggressive orchestration
configuration decreased full-set Level-1 accuracy while increasing tracebacks,
timeout mentions, tool-failure mentions, and cost estimates. The follow-up
diagnostic produced an observed one-task full-set improvement under the
recorded controls, but the magnitude is too small to support a broad capability
claim. Together, the results support a reliability-centered evaluation protocol
in which agent changes must pass both accuracy and operational-noise gates
before being treated as progress.

\section{Related Work}

\subsection{Reasoning, acting, and tool use}

ReAct established a simple but influential pattern for interleaving reasoning
traces and task-specific actions, allowing the model to update plans while
interacting with external information sources \cite{yao2023react}. Toolformer
showed that language models can be trained to decide when and how to call
external APIs, framing tool use as a learned language-model behavior rather
than a fixed wrapper around generation \cite{schick2023toolformer}. These
systems motivate tool-augmented agents, but they do not by themselves resolve
the operational question studied here: when does extra tool use or planning
increase reliability, and when does it merely amplify noise?

\subsection{Agent benchmarks and realistic environments}

GAIA evaluates general assistant behavior across search, file handling,
multimodal evidence, and concise answer synthesis \cite{mialon2023gaia}.
AgentBench broadened the evaluation of language models as agents across
multiple interactive environments \cite{liu2023agentbench}. WebArena and
WorkArena moved evaluation toward realistic browser and enterprise-software
tasks \cite{zhou2023webarena,drouin2024workarena}, while SWE-bench evaluated
software-engineering agents on real GitHub issue resolution
\cite{jimenez2023swebench}. OSWorld further emphasized open-ended interaction
with real computer environments \cite{xie2024osworld}. Tau-bench added dynamic
tool-agent-user interaction with domain-specific APIs and policies
\cite{yao2024taubench}. Relative to these benchmarks, ChromaFlow's contribution
is not a new public task suite; it is an operational analysis of a GAIA
configuration change that looked promising in smoke tests but failed under a
full Level-1 comparison.

\subsection{Orchestration and evaluation reliability}

The 2025--2026 agent literature increasingly treats orchestration as a
measurable intervention rather than an unqualified good. OrchestrationBench
explicitly targets LLM-driven planning and tool use across multi-domain
scenarios \cite{ahn2026orchestrationbench}. Recent systematization and
evaluation work argues that agentic behavior extends beyond isolated tool calls
and raises reliability, contamination, evolution, governance, and security
questions \cite{jiang2026sok,dong2026agency}. This paper provides a concrete
case study supporting that view: the recovery configuration added planning and
execution surface area, but the full-run result degraded accuracy while
increasing operational noise.

\section{System Overview}

ChromaFlow is organized around a supervisory execution controller that selects
an execution strategy, routes tool calls, monitors failure signals, and
synthesizes final answers. The controller may use direct answer synthesis for
simple tasks, or it may allocate work across specialized execution paths for
retrieval, document analysis, browser interaction, Python execution, shell
execution, and verification.

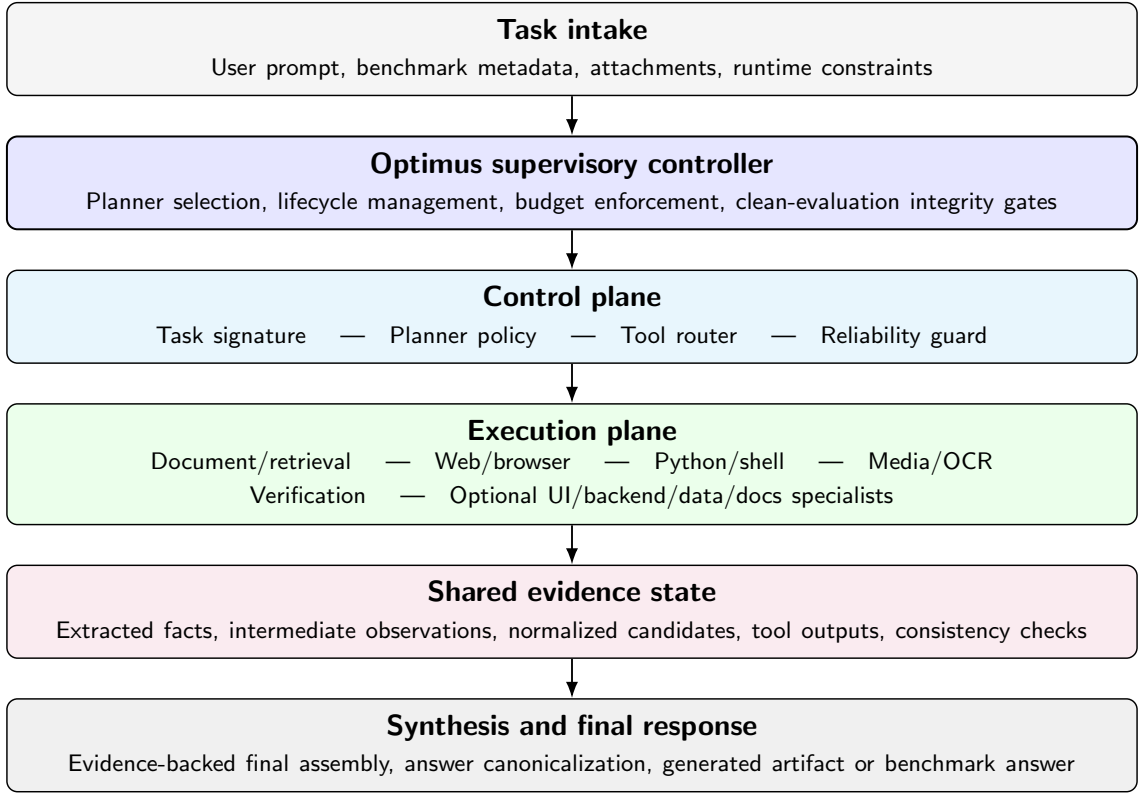
\begin{figure}[t]
\centering
\begin{tikzpicture}[
  font=\sffamily,
  >=Latex,
  stage/.style={
    draw,
    rounded corners=4pt,
    align=center,
    text width=0.88\textwidth,
    minimum height=1.0cm,
    inner sep=6pt,
    line width=0.55pt
  },
  intake/.style={stage, fill=gray!8},
  supervisor/.style={stage, fill=blue!10, line width=0.75pt},
  control/.style={stage, fill=cyan!8},
  execution/.style={stage, fill=green!8},
  state/.style={stage, fill=purple!8},
  final/.style={stage, fill=gray!12},
  arrow/.style={->, line width=0.7pt}
]

\node[intake] (input) at (0,0) {
  \textbf{Task intake}\\
  \footnotesize User prompt, benchmark metadata, attachments, runtime constraints
};
\node[supervisor, below=0.52cm of input] (optimus) {
  \textbf{Optimus supervisory controller}\\
  \footnotesize Planner selection, lifecycle management, budget enforcement, clean-evaluation integrity gates
};
\node[control, below=0.52cm of optimus] (control) {
  \textbf{Control plane}\\
  \footnotesize Task signature \quad|\quad Planner policy \quad|\quad Tool router \quad|\quad Reliability guard
};
\node[execution, below=0.52cm of control] (execution) {
  \textbf{Execution plane}\\
  \footnotesize Document/retrieval \quad|\quad Web/browser \quad|\quad Python/shell \quad|\quad Media/OCR\\[-1pt]
  \footnotesize Verification \quad|\quad Optional UI/backend/data/docs specialists
};
\node[state, below=0.52cm of execution] (evidence) {
  \textbf{Shared evidence state}\\
  \footnotesize Extracted facts, intermediate observations, normalized candidates, tool outputs, consistency checks
};
\node[final, below=0.52cm of evidence] (synthesis) {
  \textbf{Synthesis and final response}\\
  \footnotesize Evidence-backed final assembly, answer canonicalization, generated artifact or benchmark answer
};

\foreach \a/\b in {input/optimus,optimus/control,control/execution,execution/evidence,evidence/synthesis} {
  \draw[arrow] (\a.south) -- (\b.north);
}

\end{tikzpicture}%
\caption{Professionalized ChromaFlow system topology. The control plane
profiles the task, chooses an execution policy, routes tools, and enforces
reliability gates. The execution plane writes observations into a shared
evidence state before synthesis produces the final response. The layered layout
keeps routing explicit without overlapping edges or labels.}
\label{fig:architecture}
\end{figure}

Figure~\ref{fig:architecture} summarizes the system topology. The diagram uses
a separated control plane, execution plane, evidence layer, and synthesis layer
to emphasize that ChromaFlow's reliability policy is not a single component; it
is a constraint applied across routing, tool execution, and final-answer
assembly.

\subsection{Planner-directed execution}

Incoming tasks are mapped to a task signature that estimates modality,
attachment requirements, likely answer shape, and tool needs. The planner then
chooses an execution path and a tool budget. The goal is to avoid treating every
question as a broad web-research problem. Tasks with documents, spreadsheets,
images, or code artifacts should first extract local evidence before escalating
to search-heavy workflows.

\subsection{Tool layer}

The tool layer includes web search, document parsing, Python execution, shell
execution, browser automation, media handling, and final answer normalization.
Tool outputs are treated as evidence objects that can be inspected by the
answering policy. The system tracks tool exceptions, timeouts, fallback
activation, and retry behavior because those events often predict answer
quality and cost.

\subsection{Reliability layer}

The reliability layer is responsible for bounding tool calls, detecting retry
storms, handling timeouts, and preventing unavailable optional providers from
generating repeated tracebacks. It also records missing finals, direct-solver
integrity hits, elapsed time, attempts, and aggregate cost estimates. These
signals are used as run gates: a configuration should not graduate from smoke
testing to a full run unless it improves accuracy without materially increasing
operational noise.

\section{Evaluation Protocol}

The evaluation used GAIA 2023 Level-1 validation tasks. To preserve benchmark
integrity, public reporting excludes task text, task identifiers, gold answers,
model predictions, attachment names, raw URLs, and raw execution traces. Only
aggregate accuracy, aggregate operational metrics, and non-identifying
transition counts are reported.

The clean evaluation constraints were:
\begin{itemize}[leftmargin=*]
  \item direct solver paths disabled;
  \item no task-specific answer patches or task-ID routing;
  \item no publication of benchmark task text or gold answers;
  \item zero tolerance for missing final answers in completed full runs;
  \item run directories frozen for later audit;
  \item full-run launch gated by randomized smoke-run performance and noise.
\end{itemize}

The primary comparison is between a frozen full Level-1 baseline and a later
full Level-1 recovery configuration. Both runs covered 53 Level-1 validation
tasks. Additional 20-task randomized smoke runs were used as diagnostic gates,
not as proof of full-set performance.

\section{Results}

\subsection{Full Level-1 comparison}

Table~\ref{tab:full-l1} summarizes the full Level-1 comparison. The frozen
baseline achieved 29/53 correct answers. The recovery configuration achieved
27/53 correct answers. Both runs produced zero missing finals.

\begin{table}[h]
\centering
\caption{Full GAIA Level-1 validation comparison.}
\label{tab:full-l1}
\begin{tabular}{lcc}
\toprule
Metric & Frozen baseline & Recovery configuration \\
\midrule
Correct / total & 29 / 53 & 27 / 53 \\
Accuracy & 54.72\% & 50.94\% \\
Elapsed time & 6653.89 s & 7230.93 s \\
Average task time & 115.86 s & 124.89 s \\
Total attempts & 57 & 58 \\
Missing finals & 0 & 0 \\
Seed & 1782047163 & 1778025467 \\
\bottomrule
\end{tabular}
\end{table}

The recovery configuration therefore produced a net loss of two tasks. This is
not an improvement claim. It is evidence that the expanded orchestration
configuration failed to generalize from smaller diagnostic signals to the full
Level-1 task set.

\subsection{Operational noise and cost}

Table~\ref{tab:noise} reports aggregate operational telemetry. The recovery
configuration increased tracebacks, timeout mentions, tool-failure mentions,
and campaign-log cost estimates.

\begin{table}[h]
\centering
\caption{Operational noise and campaign-log cost estimates. Cost estimates are
operational estimates, not provider billing statements.}
\label{tab:noise}
\begin{tabular}{lcc}
\toprule
Metric & Frozen baseline & Recovery configuration \\
\midrule
Tracebacks & 62 & 141 \\
Timeout mentions & 769 & 939 \\
Tool-failure mentions & 170 & 282 \\
Priority cost estimate & \$197.32 & \$231.55 \\
Standard-equivalent estimate & \$98.66 & \$115.78 \\
\bottomrule
\end{tabular}
\end{table}

The important point is not merely that the recovery run cost more. The accuracy
declined while cost and noise increased. This pattern suggests that the recovery
changes added execution entropy without adding enough reasoning or verification
quality to compensate.

\subsection{Task-level movement}

Task-level comparison covered 53 common tasks. To preserve benchmark
confidentiality, only aggregate transition counts are reported. The recovery
configuration produced six correct-to-wrong transitions and four
wrong-to-correct transitions. Twenty-three tasks remained correct and twenty
tasks remained wrong.

\begin{table}[h]
\centering
\caption{Aggregate task-level movement from frozen baseline to recovery run.}
\label{tab:movement}
\begin{tabular}{lc}
\toprule
Transition type & Count \\
\midrule
Correct to wrong & 6 \\
Wrong to correct & 4 \\
Same correct & 23 \\
Same wrong & 20 \\
\bottomrule
\end{tabular}
\end{table}

The net movement was negative. Error analysis showed stronger regression
signals around code-and-document tasks, document-and-image tasks, numeric/date
answers, and no-attachment semantic research tasks. Those categories point
toward extraction, normalization, and evidence reconciliation issues rather than
simply insufficient retries.

\subsection{Smoke-run replication checks}

The smoke runs were used as randomized gates. They were not treated as proof
runs. The first post-patch randomized smoke achieved 12/20, or 60.00\%, with
zero missing finals and zero direct-solver hits. It was frozen as a positive
smoke but not as evidence of full-set improvement. The next randomized smoke
achieved 11/20, or 55.00\%, and had higher traceback and timeout noise.

\begin{table}[h]
\centering
\caption{Randomized 20-task Level-1 smoke checks.}
\label{tab:smoke}
\begin{tabular}{lcc}
\toprule
Metric & Positive smoke & Subsequent smoke \\
\midrule
Correct / total & 12 / 20 & 11 / 20 \\
Accuracy & 60.00\% & 55.00\% \\
Missing finals & 0 & 0 \\
Total attempts & 24 & 23 \\
Elapsed time & 2819.61 s & 2761.30 s \\
Seed & 1778077117 & 1778082895 \\
\bottomrule
\end{tabular}
\end{table}

This sequence supports the full-run gate. The positive smoke was encouraging,
but the subsequent smoke did not clear the threshold. Therefore, the smoke
results alone did not justify an improvement claim. They did, however, motivate
a later full-set diagnostic framed in advance as failure discovery rather than
as a pre-declared success.

\section{Follow-up Strict-Provider Diagnostic}

After the negative ablation and unstable smoke checks, ChromaFlow was evaluated
in a later full GAIA 2023 Level-1 validation diagnostic under stricter provider
and policy controls. The purpose was not to validate a broad new capability
claim; it was to test whether a cleaner execution substrate could recover any
full-set accuracy while preserving benchmark-integrity constraints.

The follow-up protocol disabled direct solvers, policy auto-loading,
meta-learning during evaluation, evolution promotion, latest-healing policy
loading, task-specific reruns, and mid-run tuning. It used clean evaluation
mode, evaluation lockdown, and a search-provider health gate before scoring.
After a pre-scoring smoke attempt showed fallback-provider rate limiting, the
full diagnostic used strict Bing-primary search. This provider route was frozen
before the full run and applied globally; it was not a task-specific
intervention.

\begin{table}[h]
\centering
\caption{Full Level-1 trajectory including the strict-provider follow-up
diagnostic.}
\label{tab:followup}
\begin{tabular}{lccc}
\toprule
Run & Correct / total & Accuracy & Interpretation \\
\midrule
Frozen baseline & 29 / 53 & 54.72\% & Reference point \\
Expanded recovery ablation & 27 / 53 & 50.94\% & Accuracy down, noise up \\
Strict-provider diagnostic & 30 / 53 & 56.60\% & Accuracy up, cost up \\
\bottomrule
\end{tabular}
\end{table}

The diagnostic produced an observed one-task full-set improvement under the
recorded controls: 30/53 rather than the prior 29/53 baseline. The magnitude is
too small to support a broad capability claim. It is nevertheless a legitimate
full-set diagnostic result because it covers all 53 Level-1 validation tasks
under a frozen protocol rather than a cherry-picked smoke sample.

\begin{table}[h]
\centering
\caption{Strict-provider diagnostic controls and telemetry. Cost estimates are
token-log based and exclude virtual-machine costs, search/API side costs, and
provider billing adjustments.}
\label{tab:followup-telemetry}
\begin{tabular}{lc}
\toprule
Metric & Value \\
\midrule
Direct solvers & Disabled \\
Evaluation lockdown & Enabled \\
Policy auto-loading & Disabled \\
Meta-learning during evaluation & Disabled \\
Evolution promotion & Disabled \\
Latest-healing policy load & Disabled \\
Search route & Strict Bing primary \\
Search health gate & Passed before scoring \\
Task-specific reruns & Forbidden \\
Mid-run tuning & Forbidden \\
Missing finals & 0 \\
Total attempts & 59 \\
Elapsed time & 5438.06 s \\
Average task time & 88.0 s \\
Average orchestration entropy score & 0.1687 \\
Token-log calls & 928 \\
Input tokens logged & 11,619,201 \\
Output tokens logged & 95,921 \\
Priority cost estimate & \$185.80 \\
Standard cost estimate & \$92.90 \\
\bottomrule
\end{tabular}
\end{table}

The follow-up therefore changes the empirical story but not the paper's
discipline. The initial conclusion remains valid: adding orchestration by
default degraded the full run. The later diagnostic shows that stricter
provider and policy controls can recover a small full-set gain, but the gain
comes with a substantial token-log cost increase. The appropriate conclusion is
that reliability improved under tighter controls, while cost efficiency remains
unresolved.

\section{Failure Analysis}

The negative ablation exposed several generic failure modes.

\subsection{Retry amplification}

Retries are useful when failures are transient. They are harmful when the
underlying error is deterministic, such as an unavailable provider, a missing
file path, an execution timeout, or a command that repeatedly exits with a
nonzero status. In those cases, retrying consumes budget and increases latency
without changing the evidence available to the model.

\subsection{Timeout accumulation}

Timeout mentions were high in both full runs and increased in the recovery
configuration. Timeout accumulation is especially damaging in benchmark
settings because it reduces the number of useful tool observations per unit
cost. It can also push the agent toward final-answer synthesis under time
pressure, which increases the risk of poorly verified answers.

\subsection{Attachment and evidence routing}

Attachment-heavy tasks require local extraction before broad search escalation.
The error clusters suggest that routing should distinguish between incomplete
local evidence and a genuinely open-ended research requirement. An attachment
should not automatically trigger expensive planner escalation if deterministic
parsing can answer the question.

\subsection{Final answer normalization}

Several regressions skewed toward numeric/date answer shapes. This suggests
that answer canonicalization should be treated as a separate reliability
component. Numeric, date, unit, and list answers should be normalized and
checked against the extracted evidence before final submission.

\section{Recovery v2 Design Implications}

The negative ablation motivates a gated orchestration policy in which planner
escalation is treated as a cost-bearing intervention rather than a default
behavior. Recovery v2 should therefore reduce orchestration entropy rather than
add more reasoning depth by default. The design should cap retries per task and
tool, stop recursive planner expansion, route simple tasks through cheaper
clean-evaluation paths, require deterministic extraction before final answers,
reconcile evidence before answer synthesis, fail closed on noisy tools, and log
orchestration depth and entropy as telemetry rather than as optimization
targets.

The next comparison should be preregistered as Baseline versus Recovery v1
versus Recovery v2 under identical seeds, prompts, routing policies, tools, and
scoring code. Recovery v2 should count as successful only if accuracy is higher
than Baseline, accuracy is preserved or improved versus Recovery v1, and
tracebacks, timeouts, tool failures, attempts, token volume, and estimated cost
are lower than Recovery v1, without increasing missing finals or
baseline-relative correct-to-wrong movement. A higher accuracy number with
doubled cost, exploding attempts, more timeouts, or hidden failures should be
treated as another unstable recovery rather than as progress.

\section{Discussion}

The results support three practical conclusions.

First, small smoke-run gains are not sufficient evidence for full-set
improvement. A 20-task smoke can discover promising reliability changes, but it
can also overrepresent an easier or more compatible slice of the benchmark. A
full-set run should require both accuracy improvement and lower operational
noise.

Second, autonomy is not the same as unbounded orchestration. An agent that
spawns more planning, retries more tools, and searches more broadly can become
less reliable. Autonomy should include the ability to stop, choose a simpler
path, and declare a tool failure non-retryable when further calls are unlikely
to help.

Third, negative ablations should be preserved. They prevent misleading progress
claims and reveal which changes increase cost without improving final answers.
For ChromaFlow, the negative ablation suggests that the next improvements should
target deterministic extraction, answer normalization, timeout policy, browser
provider fallback, and evidence-backed planner escalation before additional
full benchmark runs.

\section{Reproducibility and Artifact Policy}

The evaluation artifacts are structured as run directories containing sanitized
summaries, status files, aggregate metrics, and audit manifests. Public reports
intentionally avoid task text, gold answers, raw predictions, raw attachments,
and detailed traces. This protects benchmark integrity while still allowing
aggregate claims to be audited.

The recommended artifact policy for future public release is:
\begin{itemize}[leftmargin=*]
  \item publish aggregate metrics and non-identifying transition counts;
  \item publish code changes that are generic and not task-specific;
  \item keep private any benchmark task text, gold answers, predictions, and raw
        logs that could reveal tasks;
  \item report cost estimates as operational estimates rather than billing
        statements;
  \item label smoke runs as diagnostic unless a full-set run confirms the trend.
\end{itemize}

\section{Limitations and Threats to Validity}

This report has several limitations. The main full-run analysis is limited to
GAIA Level-1 validation tasks. It does not establish superiority across GAIA
Level-2 or Level-3, and it does not claim a top leaderboard position. The
ChromaFlow implementation also contains proprietary orchestration components
that are described at a high level rather than released in full.

The strongest threat to validity is external generalization. The result is one
framework, one benchmark level, and one 53-task validation split. It should not
be read as evidence that orchestration is generally harmful, or that a different
agent, model, provider, benchmark level, or tool stack would produce the same
movement. The claim is narrower: in this controlled ChromaFlow comparison,
expanded orchestration reduced full-set Level-1 accuracy and increased
operational noise.

The cost values are derived from campaign logs and should not be interpreted as
provider billing records. They exclude potential VM, search/API, and
provider-side adjustments. The operational-noise counters are also log-derived
and should be read as comparative telemetry rather than exact causal labels.

Another threat is measurement sensitivity. Traceback, timeout, and tool-failure
counts come from logs, so they can be affected by logging verbosity and error
wording. The paper therefore uses them as relative operational signals rather
than as exact causal measurements. Benchmark results are also sensitive to model
behavior, tool-provider availability, web conditions, and runtime policy. This
is one reason the paper emphasizes frozen protocols, audit manifests, smoke-run
gates, and negative-ablation reporting.

\section{Recommendations}

Based on the negative ablation and the later strict-provider diagnostic,
ChromaFlow should not launch Level-2 or Level-3 evaluation merely because the
follow-up run gained one Level-1 task. The next engineering work should be
generic rather than task-specific:
\begin{itemize}[leftmargin=*]
  \item reduce browser-provider fallback noise;
  \item make deterministic tool failures non-retryable;
  \item enforce process cleanup after Python and shell timeouts;
  \item improve document, spreadsheet, image, and attachment evidence routing;
  \item add numeric/date/list answer canonicalization;
  \item require randomized smoke runs and full-set diagnostics to report both
        accuracy and cost/noise tradeoffs.
\end{itemize}

Before any new full run, the benchmark protocol should be frozen in a manifest
that records the git commit, dirty diff hash, policy and prompt hashes, scorer
hash, task manifest hash, seed, model label, runtime environment, abort rules,
and comparison metrics. Individual failed tasks should not be rerun, and any
mid-run change to prompts, retries, routing, tools, extraction logic, or scoring
should invalidate the comparison.

This path preserves evaluation integrity while making the agent more useful in
practice. The goal is not to tune for a fixed task list. The goal is to reduce
generic execution failure modes that plausibly affect many tool-augmented
reasoning tasks.

\section{Conclusion}

This paper presented ChromaFlow as an operational case study in autonomous
reasoning reliability. The frozen full Level-1 baseline achieved 54.72\%, while
the expanded recovery configuration achieved 50.94\% and generated more
operational noise and higher cost estimates. The result is a negative ablation,
not an improvement claim. A later strict-provider diagnostic reached 30/53, or
56.60\%, under clean controls with direct solvers disabled, but at substantially
higher token-log cost. That follow-up is best interpreted as an observed
one-task recovery under the recorded protocol rather than as a broad capability
claim.

The broader lesson is that agentic capability must be evaluated together with
operational behavior. Planner depth, retries, tool diversity, and multi-agent
coordination can help, but they can also amplify instability. Reliable agent
systems need bounded escalation, deterministic extraction, explicit
non-retryable failure handling, evidence reconciliation, and honest run gates.
For ChromaFlow, future progress should be measured by clean full-set gains that
preserve or improve the 56.60\% strict-provider result while reducing cost,
timeouts, tracebacks, and tool failures, with no compromise to benchmark
integrity.

\end{document}